\documentclass[letterpaper, 10 pt, conference]{ieeeconf}  

\IEEEoverridecommandlockouts                              

\usepackage{graphics}
\usepackage{amsmath,amssymb}
\usepackage[table]{xcolor} 
\usepackage{booktabs}
\usepackage{subcaption}               
\usepackage{siunitx}                  
\usepackage[textsize=tiny]{todonotes} 
\usepackage{lipsum}                   
\usepackage{dblfloatfix}              
\usepackage{scalerel}
\usepackage{tikz, pgfplots}
  \usetikzlibrary{svg.path}

\makeatletter
\let\NAT@parse\undefined
\makeatother
\usepackage{hyperref}

\newcommand{\fig}[1]{Fig.~\ref{#1}}

\newcommand{\tab}[1]{Table~\ref{#1}}

\newcommand\copyrighttext{%
    \footnotesize \copyright{ }2023 IEEE. Personal use of this material is permitted. Permission from IEEE must be obtained for all other uses, in any current or future media, including reprinting/republishing this material for advertising or promotional purposes, creating new collective works, for resale or redistribution to servers or lists, or reuse of any copyrighted component of this work in other works.}
\newcommand\copyrightnotice{%
    \begin{tikzpicture}[remember picture,overlay]
    \node[anchor=south,yshift=15pt,xshift=0pt] at (current page.south) {\parbox{\dimexpr\textwidth-\fboxsep-\fboxrule\relax}{\copyrighttext}};
    \end{tikzpicture}%
}

\definecolor{orcidlogocol}{HTML}{A6CE39}
\tikzset{
  orcidlogo/.pic={
    \fill[orcidlogocol] svg{M256,128c0,70.7-57.3,128-128,128C57.3,256,0,198.7,0,128C0,57.3,57.3,0,128,0C198.7,0,256,57.3,256,128z};
    \fill[white] svg{M86.3,186.2H70.9V79.1h15.4v48.4V186.2z}
                 svg{M108.9,79.1h41.6c39.6,0,57,28.3,57,53.6c0,27.5-21.5,53.6-56.8,53.6h-41.8V79.1z M124.3,172.4h24.5c34.9,0,42.9-26.5,42.9-39.7c0-21.5-13.7-39.7-43.7-39.7h-23.7V172.4z}
                 svg{M88.7,56.8c0,5.5-4.5,10.1-10.1,10.1c-5.6,0-10.1-4.6-10.1-10.1c0-5.6,4.5-10.1,10.1-10.1C84.2,46.7,88.7,51.3,88.7,56.8z};
  }
}
\newcommand\orcidicon[1]{\href{https://orcid.org/#1}{\mbox{\scalerel*{
\begin{tikzpicture}[yscale=-1,transform shape]
\pic{orcidlogo};
\end{tikzpicture}
}{|}}}}

\usepackage{url}

\title{\LARGE \bf
  RobotKube: Orchestrating Large-Scale Cooperative Multi-Robot Systems with Kubernetes and ROS
}

\author{
  Bastian Lampe\textsuperscript{\orcidicon{0000-0002-4414-6947}}*, 
  Lennart Reiher\textsuperscript{\orcidicon{0000-0002-7309-164X}}*, 
  Lukas Zanger\textsuperscript{\orcidicon{0009-0005-0292-2660}}*, 
  Timo Woopen\textsuperscript{\orcidicon{0000-0002-7177-181X}}*, \\
  Raphael van Kempen\textsuperscript{\orcidicon{0000-0001-5017-7494}}, 
  and Lutz Eckstein
  \thanks{*These authors contributed equally to this work. \newline All authors are with the Institute for Automotive Engineering~(ika), RWTH Aachen University, 52074 Aachen, Germany. {\tt\small \{firstname.lastname\}@ika.rwth-aachen.de}}
}

\begin{document}

\bstctlcite{IEEEexample:BSTcontrol}
\maketitle
\thispagestyle{empty}
\pagestyle{empty}
\copyrightnotice

\begin{abstract}
  Modern cyber-physical systems (CPS) such as Cooperative Intelligent Transport Systems (C-ITS) are increasingly defined by the software which operates these systems. In practice, microservice architectures can be employed, which may consist of containerized microservices running in a cluster comprised of robots and supporting infrastructure. These microservices need to be orchestrated dynamically according to ever changing requirements posed at the system. Additionally, these systems are embedded in DevOps processes aiming at continually updating and upgrading both the capabilities of CPS components and of the system as a whole. In this paper, we present RobotKube, an approach to orchestrating containerized microservices for large-scale cooperative multi-robot CPS based on Kubernetes. We describe how to automate the orchestration of software across a CPS, and include the possibility to monitor and selectively store relevant accruing data. In this context, we present two main components of such a system: an event detector capable of, e.g., requesting the deployment of additional applications, and an application manager capable of automatically configuring the required changes in the Kubernetes cluster. By combining the widely adopted Kubernetes platform with the Robot Operating System (ROS), we enable the use of standard tools and practices for developing, deploying, scaling, and monitoring microservices in C-ITS. We demonstrate and evaluate RobotKube in an exemplary and reproducible use case that we make publicly available at \href{https://github.com/ika-rwth-aachen/robotkube}{\nolinkurl{github.com/ika-rwth-aachen/robotkube}}.
\end{abstract}

\section{Introduction}

Cyber-physical systems (CPS), such as Cooperative Intelligent Transport Systems (C-ITS), comprise a diverse range of interconnected entities that can vary in number and type. Next to automated vehicles, there may exist traffic control systems, road side units, control centers, and (edge-) clouds providing additional services. This leads to the question of how the software for all these different systems can work together efficiently and safely over time.

One popular approach to build such complex software systems are microservice architectures~\cite{Nadareishvili_Mitra_McLarty_Amundsen_2016}. The complete system is broken up into many loosely coupled, fine-grained services. They communicate through predefined protocols. Services in this architecture are typically run in containers, a type of virtualization allowing services to be run in isolation of each other.
A major challenge when using containers in microservice architectures is the orchestration of said containers. Orchestration involves the automated deployment, scaling, and management of containerized applications. Kubernetes is a popular open source system for this task used by many large software companies worldwide. It already comes with a lot of the capabilities that would also be needed in a C-ITS employing a microservice architecture.~\cite{kubernetes}

Kubernetes lacks methods for orchestration that are domain-specific, e.g., to C-ITS. Specific tasks that are needed only at certain times, like deploying required applications, or the recording of relevant data, must be defined either by C-ITS administrators, or programmatically by C-ITS developers. The required tasks may depend on the specific content of data exchanged in the Kubernetes cluster instead of on metadata, e.g., the load of a server. RobotKube describes new software components, which are themselves containerized and can be orchestrated by Kubernetes, that extend the regular capabilities of Kubernetes to those specific to robotic systems in general and to C-ITS in particular.

We present an \textit{event detector}, a software component based on the Robot Operating System (ROS), that can take as input any data provided by services in the cluster, analyze the data based on analysis rules implemented by developers, and formulate tasks based on the result of these analyses. Possible tasks include deploying new applications, the reconfiguration of existing applications, or the recording of a set of data which was determined relevant for further analysis and storage as part of a DevOps process.

In addition, we introduce an \textit{application manager} acting as the link between the event detector and the Kubernetes control plane. It translates requirements for specific applications or configurations communicated by the event detector into a specific workload for the Kubernetes cluster. Together with the event detector and its operator plugin, the application manager forms an \textit{operator application} that automates the management of the cluster based on occurring events.

We apply RobotKube in an exemplary use case where a cloud-based operator application detects when two automated vehicles approach each other. It then deploys communication software allowing them to transmit additional sensor data to the cloud. There, a dynamically deployed \textit{recording application} gathers corresponding sensor and location data and saves them to a database. This use-case represents the first step in a data-driven DevOps process allowing Collective Learning~\cite{Lampe2022}. The location data of one vehicle can serve as a label in the sensor data set of the the other vehicle and vice versa. Training data sets for supervised learning can hereby be generated automatically without human labeling. 

This use case is of course only one of many which are possible with the approach described in this work.

In summary, our work makes these main contributions:

\begin{itemize}
  \item Introduction of RobotKube, an approach to automatically orchestrating containerized microservices for large-scale cooperative multi-robot systems based on Kubernetes and ROS.
  \item Presentation of two main components of RobotKube: an event detector acting upon the occurrence of data patterns, e.g., requesting the deployment of additional applications; and an application manager capable of automatically configuring required changes in the Kubernetes cluster. Together, the two components act as an automated Kubernetes operator application.
  \item Examination of RobotKube in an exemplary use case involving the automated deployment of various software components to multiple connected C-ITS nodes, and the recording of data relevant to the use case.
  \item Publication of the experimental setup and involved Docker images to make the exemplary use case reproducible and allow other researchers to see the system in action.
\end{itemize}

\section{Related Work}

Our work heavily builds upon containerization, Kubernetes, and the Robot Operating System (ROS). For an introduction to these tools, see~\cite{kubernetes, Burns2019-ls, ros, Kane2023-zd}. Our approach combines advantages from each of these tools.

\textit{Containerization}, the process of encapsulating applications and their dependencies into isolated units, offers a range of benefits for application management. It enables rapid deployment and over-the-air updates. Compatibility issues are reduced, and applications become portable across machines. Components can be easily reused and shared with others. Containers have a lightweight footprint and minimal overhead, optimizing resource utilization. Maintenance is simplified as applications and dependencies are encapsulated. This makes them suitable to container orchestration software~\cite{Kane2023-zd}. Different tools for containerization exist. The authors in \cite{CanonicalLtd_KeepEnterpriseROS_2020, CanonicalLtd_RosDocker_2023, CanonicalLtd_SnappingOutofDocker_2023} propose \textit{snaps} for production robots because they make available interfaces for accessing low-level hardware and come with a robust update system. In our approach, we choose \textit{Docker containers} because of their popularity, their easy integration into Kubernetes, and their versatility in the development phase of our approach. It is conceivable to move to a different containerization framework later.

\textit{Orchestration} is "the automated configuration, management, and coordination of computer systems, applications, and services"  and "helps to more easily manage complex tasks and workflows"~\cite{Orchestration}. \textit{Kubernetes} is one popular orchestration platform that brings several advantages to the deployment and management of containerized applications. It abstracts infrastructure complexities, ensuring portability across different environments. It enables automatically adjusting resources based on workload demands, and supports horizontal scaling. Kubernetes offers self-healing capabilities, automatically restarting failed containers, and supports rolling updates and rollbacks for seamless application upgrades. It provides service discovery and load balancing mechanisms for efficient traffic routing and distribution. Kubernetes' declarative configuration allows for reproducibility and version control, reducing configuration drift.~\cite{kubernetes, Burns2019-ls}. 

The described capabilities make Kubernetes suitable to large-scale robotic systems. Popular alternatives include \textit{docker~compose}~\cite{DockerCompose} and \textit{docker swarm}~\cite{DockerSwarm}. In the context of robotics, Kubernetes is employed for Industry 4.0 applications~\cite{Barletta2022}. Examples in the context of C-ITS include~\cite{Microsoft_MercedesBenzCreatesContainerdriven_2020, Slamnik_2021, ros_containerization_2020, DECICE_2023}. The current trend towards software-defined vehicles reinforces the importance of orchestration tools like Kubernetes in C-ITS. Their use can reduce the time period for the release of software updates~\cite{Microsoft_MercedesBenzCreatesContainerdriven_2020, over_the_air_update_framework_2022}. An important driver of the need for orchestration software can also be found in the increased connectedness between automated vehicles and supporting infrastructure. Increased research activity is found regarding the use of sensored infrastructure and edge clouds for C-ITS. Sensored road side units can play an important role in supporting automated vehicles in their operation~\cite{kloeker_2020}, e.g., by providing additional perception data to mitigate challenges like occlusions and limited sensor range~\cite{V2I_environment_perception_2021}. Edge clouds or clouds can be used for function offloading to make use of more powerful compute resources outside of vehicles~\cite{edge_cloud_computing_ITS_2022}. An orchestration system managing C-ITS shall therefore encompass all interacting subsystems including automated vehicles and supporting infrastructure. While other research focuses on, e.g., real-time capabilities of orchestration~\cite{Barletta2022}, and requires human operators~\cite{ichnowski2023fogros2}, our work describes an approach to automate cluster operations. In contrast to other automated systems in which operators are either only cloud-based \cite{Slamnik_2021}, or vehicle-based~\cite{kampmann_asoa_2019}, RobotKube is in principle decentralized and agnostic to whether operators are, e.g., based in the cloud, in sensored road side units, or in automated vehicles.

\textit{Applications and Services} for robotic systems are often developed using common software frameworks and middlewares. For RobotKube, we choose \textit{ROS}, a widely adopted set of open source software libraries and tools for building robot applications. It provides advantages for development through its ecosystem of tools, libraries, and capabilities. Its global community continuously enhances the software ecosystem. ROS is widely used in research projects and production robots worldwide. It supports various platforms, including Linux, Windows, and microcontrollers. ROS is open source, offering customization and integration flexibility~\cite{ros}. 

Alternatives to ROS include YARP~\cite{YARP}, EB~Assist ADTF~\cite{ADTF}, ASOA~\cite{kampmann_asoa_2022}, and AUTOSAR Adaptive~\cite{AUTOSARadaptive}. These come with different capabilities and objectives, but can partly also be made compatible with ROS. There also exist the products of Apex.AI~\cite{Becker2022} which build upon ROS.

\begin{figure*}[t]
  \centering
  \includegraphics[clip, trim=5.31cm 6.8cm 5.31cm 6.6cm, width=\linewidth]{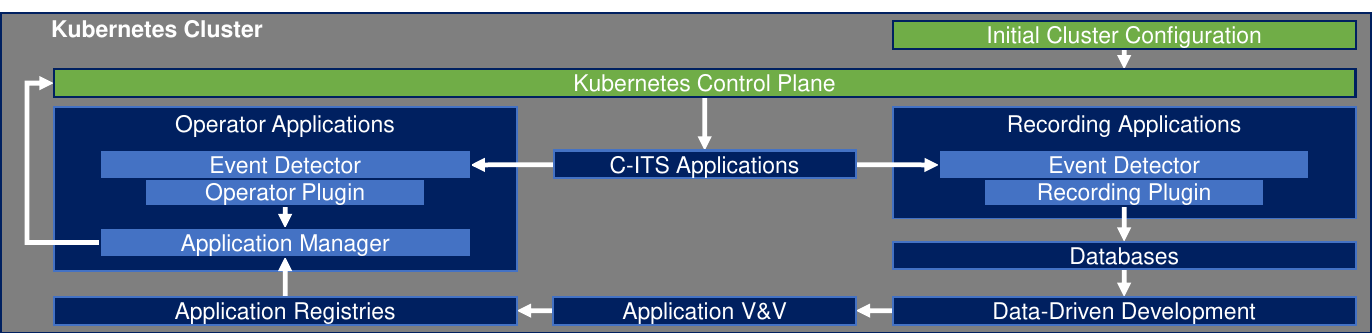}
  \caption{The overall system architecture of the RobotKube approach contains its essential components and depicts the general orchestration and application development process in a C-ITS. An initial configuration is used to deploy C-ITS applications. These include two types of special applications. Operator applications can interact with the control plane to, e.g., deploy additional applications, which are made available in application registries. Recording applications can selectively store data in databases. The stored data can be used for data-driven development of new applications, or for application updates. All applications need to be verified and validated before they are made available in the application registries.}
  \label{fig:architecture}
\end{figure*}

Due to their extensive capabilities, a combination of our chosen tools is already used in various contexts of robotic applications in general and C-ITS in particular. Based on their characteristics, they are especially suited to be employed in microservice or service-oriented architectures~(SOA). Automotive SOAs that make use of these tools have been proposed in various initiatives both in academia and in industry~\cite{kampmann_asoa_2019, kampmann_asoa_2022, FurstBechter_AUTOSARConnectedAutonomous_2016, WoopenEtAl_UNICARagilDisruptiveModular_2018a, RumezEtAl_OverviewAutomotiveServiceOriented_2020, PohnlEtAl_MiddlewareJourneyMicrocontrollers_2022, ScalableOpenArchitecture}.

\section{RobotKube}\label{sec:RobotKube}

RobotKube is an approach to orchestrating containerized ROS-based robotic applications in microservice architectures using Kubernetes. We present and apply our approach in the context of C-ITS, but also point to its applicability to large-scale robotic systems in general. RobotKube describes new software components in a novel architecture that enables a highly automated operation of C-ITS, including dynamic software deployments, data-driven development, and verification and validation of applications.

Building on top of Kubernetes gives access to many features relevant for the operation of a C-ITS. These include, but are not limited to: fault tolerance through high availability nodes, sophisticated rollout and rollback processes, self-healing mechanisms, load balancing mechanisms, and on-demand horizontal autoscaling. The incorporation of ROS similarly enables access to and usage of a vast existing ecosystem of open source software and tools for robotics applications, including automated vehicles and C-ITS.

\subsection{System Architecture}

\begin{table*}[t]
  \smallskip
  \smallskip
  \caption{Design principles for the RobotKube approach, including specific design principles regarding operator applications}
  \label{tab:design_principles}
  \renewcommand{\arraystretch}{1.5}
  \centering
  \begin{tabular}{p{0.4\linewidth} p{0.55\linewidth}}
    \toprule
    \textbf{Design Principle} & \textbf{Example} \\
    \midrule

    Connected agents make at least some of their \textit{compute units} available as nodes to a Kubernetes cluster. & A C-ITS cluster consists of nodes in the cloud, on edge servers, in control centers, in sensored road side units, and in automated vehicles. \\

    An \textit{application} is a set of one or several microservices with a particular purpose. & An environment perception application may be composed of two separate microservices for object detection and object tracking. \\

    An \textit{operator} is a person or software that manages application deployments, life cycles, cluster configurations and more. & A cloud-based operator application detects a vehicle approaching an intersection and deploys a perception application to a road-side unit to support the vehicle. \\
    
    Different applications are managed by different \textit{decentralized application-specific} operators. & In addition to the above cloud-based operator application for deploying supportive infrastructure functions, the deployment of a recording application is handled by another, separate vehicle-based operator application. \\

    \midrule

    Each \textit{microservice} is packaged into its own container. & Two microservices of an environment perception application, object detection and object tracking, are separately packaged into their own containers. \\

    Applications are in principle \textit{node agnostic}. & If one node is unavailable, an application may be deployed to a different node capable of running the application.  \\

    Application \textit{updates} are conducted by updating the individual container images of an application. & The object detection microservice of an environment perception application is updated independently of other containers of that application. \\

    \midrule

    Operator applications may deploy other operator applications forming \textit{operator application chains}. & One cloud-based operator application deploys another vehicle-based operator application to vehicles approaching an intersection. The vehicle-based operator application detects uncertainty in order to eventually deploy an environment perception application on the connected intersection infrastructure. \\

    Operator applications \textit{select} adequate applications and their components from \textit{application registries}. & The application manager of an operator application for environment perception chooses adequate object detection and object tracking services based on requirements defined in the task description received from the event detector, and based on guarantees associated with applications in the \textit{application registries}. \\

    Operator applications support \textit{automatic conflict resolution}. & If there are no compute resources left in a vehicle, an operator application is able to cancel or postpone deployments, or resolve the conflict by other means such as offloading to a different connected agent. \\

    \bottomrule
  \end{tabular}
  \renewcommand{\arraystretch}{1.0}
\end{table*}

The overall system architecture of RobotKube is depicted in Fig.~\ref{fig:architecture}. A manually defined \textit{initial cluster configuration} in the form of a set of \textit{C-ITS applications} is deployed by the \textit{Kubernetes control plane} to all agents that are part of the cluster and part of the C-ITS. Initially, the C-ITS administrators act as operators who define the cluster configuration and the set of initial deployments.
\begin{itemize}
  \item An \textit{application} within the scope of \textit{RobotKube} is a set of one or several microservices with a particular purpose within the C-ITS.
  \item An \textit{operator} within the scope of \textit{RobotKube} is a person or software that manages application deployments, life cycles, cluster configurations and additional cluster management tasks within the C-ITS.
\end{itemize}
Applications are developed as part of a \textit{data-driven development} process. Before being deployed, each application and its services must pass \textit{application V\&V}, i.e., the verification and validation of their desired functioning. Applications that have passed this stage are made available in \textit{application registries}.

A core goal of RobotKube is to automate the cluster operation through \textit{operator applications} that can automatically deploy, configure, and manage applications in the cluster. The operator applications act on developer-defined events in the cluster. \textit{Events} are associated with occurences of certain patterns in the data exchanged in the cluster, and are detected by dedicated \textit{event detector} components. Based on the events, a second component, the \textit{application manager}, issues new deployments or reconfigurations. As an example, an operator application may detect that a connected vehicle is approaching an intersection and then automatically deploys a supportive function onto a nearby sensored road-side unit.

The detection of developer-defined events is also relevant for \textit{recording applications} that enable on-demand data recording to a database. Recorded data can in turn be used for data-driven development of C-ITS applications, including automated collective learning techniques~\cite{Lampe2022}.

\tab{tab:design_principles} lists more key design principles of the RobotKube approach as a whole as well as design principles specific to operator applications in RobotKube. Note that the list of design principles is neither exhaustive nor set in stone, but acts a a guiding foundation for the presented approach and its individual components. We plan to continually revisit and refine the design principles in future work.

\subsection{Event Detector}

The event detector is an integral part both of operator applications and of recording applications. Its main purpose is to detect and act upon developer-defined events that are associated with patterns in the data exchanged in the cluster. An event detector is composed of three main subcomponents:

\textit{Buffer} -- All incoming data in the form of ROS messages is buffered for a configurable amount of time. It is infeasible and also undesirable to permanently store all accruing data, but the detection of events may require access to a history of data. The buffer is realized as a ring buffer covering a configurable duration of the immediate past. The buffer and the event detector in general are agnostic to specific data types and therefore compatible with all kinds of data exchanged in the cluster.

\textit{Analysis} -- The data available in the ring buffer is periodically analyzed in order to detect events. Events are associated with data patterns that are identified through an application-specific, developer-defined analysis of the buffered data. In order to support the detection of arbitrary data patterns, the event detector provides a generic and easily extensible framework for accessing the data buffer and for analyzing the data with regard to occurring events.

\textit{Action Plugin} -- Having detected an event, action plugins implement the resulting consequence. An \textit{operator plugin}, e.g., is used for requesting the deployment or reconfiguration of applications. It forwards a corresponding task description to an application manager. An event detector in combination with an operator plugin and an application manager forms an \textit{operator application}. Similarly, an event detector in combination with a recording plugin forms a \textit{recording application}. Through its action plugin mechanism, the event detector is designed to also cover use cases beyond the presented operator and recording applications.

The event detector software component is implemented as a high-performance C++ ROS node.

\subsection{Application Manager}

The application manager is an integral part of operator applications. It receives a task description from an event detector`s operator plugin and translates it to a specific Kubernetes workload definition, which is then transmitted to the Kubernetes control plane. 

The composition of requested applications from available microservices is handled by an application manager. The corresponding event detector only formulates high-level requirements. The application manager then identifies suitable containerized services, configures and links them to form the requested applications, which are then deployed to appropriate nodes in the cluster. Within the context of RobotKube, application managers are the preferred way to interact with the Kubernetes control plane in the context of launching and managing applications.

An application manager is not only capable of launching new applications, but also of managing existing applications it has launched. As part of the task description, application managers can also be requested to reconfigure existing applications, or to shut down running applications. An application manager is also responsible for deciding whether to issue requested Kubernetes workloads in the first place. Task descriptions transmitted to an application manager therefore only represent an intent or a request, not an obligation. If a requested deployment is not possible, it is the application manager's responsibility to resolve the conflict.

The application manager software component is implemented as a Python ROS node invoking the Kubernetes Python API to interact with the Kubernetes control plane.

\section{Experimental Setup}\label{sec:ExpSetup}

We apply our approach in an exemplary use case with the goal of demonstrating its main abilities, namely employing an application-specific event detector for the detection of an event to trigger the deployment of an additional application in the cluster. In particular, this involves

\begin{itemize}
  \item detecting an event in an event detector based on data that is exchanged in the cluster;
  \item transmitting a task description message containing high-level requirements regarding desired applications from the event detector to an application manager via the operator plugin;
  \item configuring the application deployments in an application manager based on the received task description;
  \item deploying the requested applications in the cluster;
  \item and managing the running applications over time.
\end{itemize}

In our exemplary use case, we aim at automatically deploying a recording application in certain situations encountered by automated vehicles with the goal to selectively gather data in a cloud-based database. The individual vehicles have insufficient information for the decision when to send the desired data to the cloud for storage. A cloud-based operator application shall therefore automatically trigger the deployment of a cloud-based recording application plus the required communication components. The data collected that way would enable collective learning~\cite{Lampe2022} methods. A detailed use case description follows.

\begin{itemize}
  \item $N$ vehicles $V_0 ,\ldots, V_N$ follow their current routes and are part of a C-ITS.
  \item $M \leq N$ vehicles $V_0 ,\ldots, V_M$ are equipped with lidar sensors producing point clouds.
  \item All vehicles send their current location/pose at a frequency of $f_p$ to a cloud server $C$, where an event detector with operator plugin receives the data. In order to save bandwidth, no lidar point clouds are sent initially.
  \item The cloud-based event detector with operator plugin continually analyzes the vehicles' poses in order to detect when any two of the lidar-equipped vehicles $V_0 ,\ldots, V_M$ are within a distance of $d_{start}$ to each other.
  \item Once any two lidar-equipped vehicles $V_i, V_j, i \neq j, i \leq M, j \leq M$ are close enough to each other, the event detector with operator plugin issues a request to launch a recording application for recording the two vehicles' poses and point clouds.
  \item A cloud-based application manager receives the request. To realize the requested application, it starts communication modules and a recording application in the cloud. The added communication modules transmit the point clouds at a frequency of $f_{pc}$ from $V_i$ and $V_j$ to $C$.
  \item The event detector with recording plugin is configured to store poses and point clouds from $V_i$ and $V_j$ in a database without further analysis.
  \item Once vehicles $V_i$ and $V_j$ have veered away from each other by more than a distance of $d_{stop}$, the deployed recording application is shut down, including the point cloud transmission.
\end{itemize}

\begin{figure*}[t]
  \centering
  \includegraphics[clip, trim=1.85cm 3.65cm 1.85cm -0.27cm, width=\linewidth]{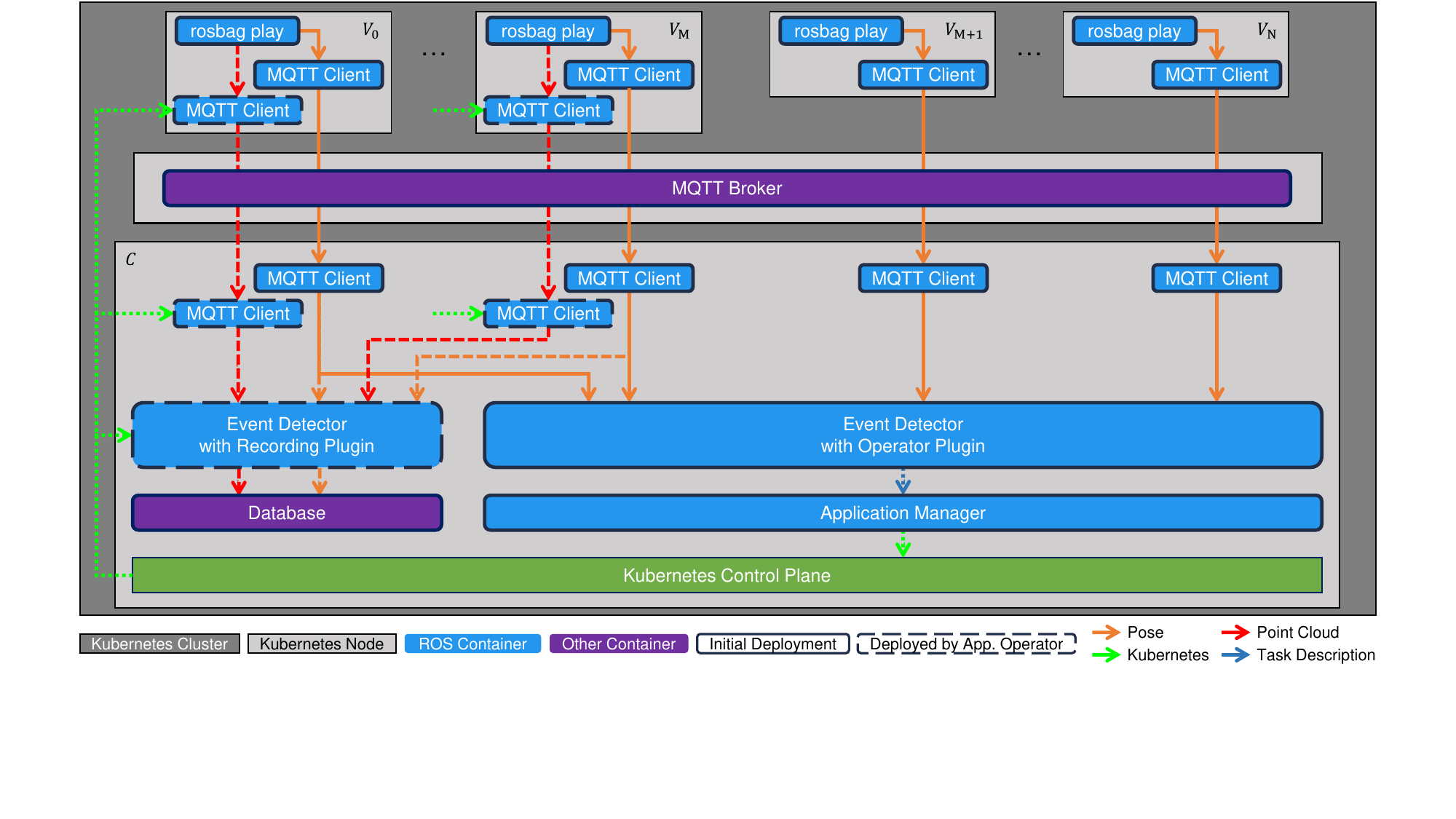}
  \caption{Software components involved in experimental setup: vehicle poses (orange lines) are sent from vehicles $V_0 ,\ldots, V_N$ to the cloud $C$ using MQTT; an event detector with operator plugin triggers a task description (blue line), when two lidar-equipped vehicles are close to each other; the task description, asking to record pose and point cloud data of the two involved vehicles $V_0, V_M$, is received and processed by an application manager; the application manager transmits a workload request to the Kubernetes Control Plane in order to launch new communication modules and a recording application (dashed lines and blocks); poses and point clouds are stored in a database.}
  \label{fig:experimental_setup}
\end{figure*}

The concrete configuration of experimental parameters is found in \tab{tab:parameters}. The involved cluster components and data flows are illustrated in \fig{fig:experimental_setup}. Individual software components are deployed as Docker containers in Kubernetes pods, running on Kubernetes nodes. Most software components are ROS-based. We simulate live vehicle data by playing back ROS bags recorded in simulation. Within the scope of one Kubernetes node, data is exchanged in the form of ROS messages. For the data transmission from vehicles to cloud, a brokered communication model using \textit{MQTT} is chosen. Dedicated \textit{mqtt\_client} ROS nodes bridge ROS messages to MQTT and vice-versa~\cite{reiher_2022}. Note that other communication models could be employed and that the communication latencies within the single-host \textit{KinD} cluster do not realistically model real-world conditions. Recording applications store ROS message data in a \textit{MongoDB} database.

\begin{table}[b]
  \caption{Configuration of experimental parameters}
  \label{tab:parameters}
  \centering
  \begin{tabular}{clr}
    \toprule
    \textbf{Parameter} & \textbf{Description} & \textbf{Value} \\
    \midrule
    $N$       & Number of vehicles                      & 15      \\
    $M$       & Number of lidar-equipped vehicles       & 2       \\
    $f_p$     & Pose frequency                          & 100 Hz  \\
    $f_{pc}$  & Point cloud frequency                   & 10 Hz   \\
    $d_{start}$ & Trigger distance between $V_i$, $V_j$   & 400 m   \\
    $d_{stop}$ & Stopping distance between $V_i$, $V_j$  & 500 m   \\
    \bottomrule
  \end{tabular}
\end{table}

In order to simulate and test the presented use case, we set up a Kubernetes cluster using \textit{Kubernetes-in-Docker~(KinD)}~\cite{kind}. This setup allows to run a multi-node Kubernetes cluster in a controlled environment on a single machine, which also enhances reproducibility for other researchers.

\section{Evaluation}\label{sect:ResultsAndDiscussion}

The previously described experimental setup allows us to test and evaluate the applicability of our approach for complex C-ITS use cases such as the one at hand. The setup mainly serves the purpose of demonstrating RobotKube's capabilities and giving an idea of how the approach translates into practice.

\fig{fig:experimental_setup} illustrates the quickly growing complexity of seemingly simple use cases for C-ITS and large-scale CPS in general. In a distributed microservice architecture, the number of software components quickly outgrows the number of connected hardware components. Combined with the dynamic nature of C-ITS, an efficient and manageable orchestration approach becomes a key enabler.

RobotKube's core orchestration component, i.e., operator applications, allow a Kubernetes-orchestrated cluster to dynamically act upon data that is currently being exchanged in the cluster. Using event detectors and application managers as described by RobotKube, we can successfully demonstrate a fully-automated event-based distributed data collection use case in a C-ITS. Since the experimental setup is made open source as part of this work, the exemplary use case and the accompanying cluster behavior can also be reproduced and studied by other researchers.

As a first quantitative evaluation of the presented approach, we describe different involved latencies. It is important to note that the results depend on the individual hardware setup and use case. Therefore, their main purpose is to give a general idea of the involved latencies, but they cannot be generalized easily. Thus, only approximate values are given here. Nonetheless, considering the involved latencies remains crucial in the development of safe C-ITS.
\begin{enumerate}
  \item \textit{Communication:} Communication latencies play a large role in all distributed systems, especially if wireless communication is involved as required in C-ITS. Concrete latencies depend on various factors such as the used communication technology, e.g., 5G or ITS-G5. Note that communication latencies are largely neglected in our experimental use case, as the distributed cluster of nodes is only virtualized.
  \item \textit{Event detection:} The developer-defined data analysis for detecting events naturally depends on the complexity of the data analysis. Here, detecting that two vehicles are close is a matter of a few milliseconds.
  \item \textit{Translation to Kubernetes workload:} The application manager translates the event detector's task description into a matching Kubernetes workload definition. This step involves the composition of a requested application from a list of available microservices, the configuration of to-be-launched components, validity checks, and possibly additional information requests via the control plane. Overall, latencies in the range of one hundred milliseconds can be expected.
  \item \textit{Cluster reconciliation:} Having received a workload definition from the application manager, the Kubernetes control plane induces the desired cluster state through a reconciliation process. At this point, new applications in the form of Kubernetes pods and potentially other Kubernetes components are launched, reconfigured, or shut down. In our experimental use case, the reconciliation phase is responsible for the majority of the total latency: the four new MQTT clients and the event detector with recording plugin take approximately five seconds to assume operation.
  \item \textit{Data storage:} A recording application stores data to a database -- in our use case, without further analysis or event detection. The data storage latency naturally scales with the amount of data to store. Given sufficiently large data buffers, it can also run asynchronously from the other processes. In our use case, ten seconds worth of pose and point cloud data from the two involved vehicles are written to the database in approximately half a second.
\end{enumerate}

The largest share of latency observed between event and data storage is attributed to the cluster reconciliation. Launching containers in the cluster comes with an overhead in the range of several seconds. While this latency is still acceptable for many C-ITS use cases, it poses constraints on the kinds of applications that can be realized in a C-ITS operated in this way. More advanced techniques like pre-launching idle containers are expected to open up the approach to more use cases over time.

\section{Conclusion}

The presented approach aims to automate the orchestration of microservices in multi-robot CPS such as C-ITS built upon Kubernetes and ROS. For this purpose, we describe design principles, provide necessary new software components, and place them in an overall C-ITS architecture connecting dynamically deployed applications, automated data-driven development, and the verification and validation of applications running in a C-ITS. Two essential types of applications are explained in this paper: \textit{operator applications} and \textit{recording applications}.

To enable these applications, the two software components \textit{event detector} and \textit{application manager} are developed. With the event detector, developer-defined events in a Kubernetes cluster can be detected, and high-level requirements regarding new applications are defined. The application manager may then deploy or reconfigure specific applications in the cluster based on these requirements and available resources. The overall approach is demonstrated in an exemplary use case. Important latencies like the startup time of an automatically deployed application are examined. The results underline that deployments of new applications or reconfiguration need to take orchestration latencies into account.

In general, our approach benefits from the vast capabilities of Kubernetes and ROS that reach far beyond those tackled in this paper. Here, we want to lay the foundation to apply the presented approach to a wide range of use cases in robotics in general and C-ITS in particular. For this purpose, and to make our research reproducible, we publish the software, data, and Kubernetes configurations used in our experiments.

\section{Acknowledgements}

This research is accomplished within the research projects "AUTOtech.\textit{agil}"~(FKZ~1IS22088A), "UNICAR\textit{agil}" (FKZ~16EMO0284K), and "6GEM"~(FKZ~16KISK036K). We acknowledge the financial support by the Federal Ministry of Education and Research of Germany (BMBF).




\bibliographystyle{IEEEtran}
\bibliography{root}

\end{document}